\documentclass[runningheads]{llncs}
\usepackage{graphicx}
\usepackage{comment}
\usepackage{amsmath,amssymb} 
\usepackage{color}
\usepackage{amsfonts,amssymb}
\usepackage{bbm}

\def\etal{\emph{et al.}}

\def\MYFLAG{submit}
\ifdefined\MYFLAG
	
	\def\gu{black}

\else
	
	\def\gu{red}

\fi

\begin{document}
\pagestyle{headings}
\mainmatter
\def\ECCVSubNumber{2216}  

\title{Finding It at Another Side:\\ A Viewpoint-Adapted Matching Encoder for Change Captioning } 

\titlerunning{Finding It at Another Side}
%
\author{Xiangxi Shi\inst{1} \and
Xu Yang\inst{1} \and
Jiuxiang Gu\inst{2} \and Shafiq Joty\inst{1} \and Jianfei Cai\inst{1, 3}}
\authorrunning{X. Shi et al.}
%
\institute{Nanyang Technological University, 50 Nanyang Avenue, Singapore 
\\
\email{\{xxshi, srjoty, asjfcai\}@ntu.edu.sg, s170018@e.ntu.edu.sg}
\and
Adobe Research, USA\\
\email{jigu@adobe.com}
\and
Monash University, Victoria 3800, Australia\\
\email{jianfei.cai@monash.edu}}
\maketitle

\begin{abstract}

Change Captioning is a task that aims to describe the difference between images with natural language. Most existing methods treat this problem as a difference judgment without the existence of distractors, such as viewpoint changes. However, in practice, viewpoint changes happen often and can overwhelm the semantic difference to be described. In this paper, we propose a novel visual encoder to explicitly distinguish
viewpoint changes from semantic changes in the change captioning task. Moreover, we further simulate the attention preference of humans and propose a novel reinforcement learning process to fine-tune the attention directly with language evaluation rewards. Extensive experimental results show that our method outperforms the state-of-the-art approaches by a large margin in both Spot-the-Diff and CLEVR-Change datasets.

\end{abstract}

\section{Introduction}

The real world is a complex dynamic system. Changes are ubiquitous and significant in our day-to-day life. As humans, we can infer the underlying information from the detected changes in the dynamic task environments. 
For instance, a well-trained medical doctor can judge the development of a patient's condition better by comparing the CT images captured in different time steps in addition to locating the lesion.

\begin{figure}[t]
    \centering
    \includegraphics[width=\linewidth]{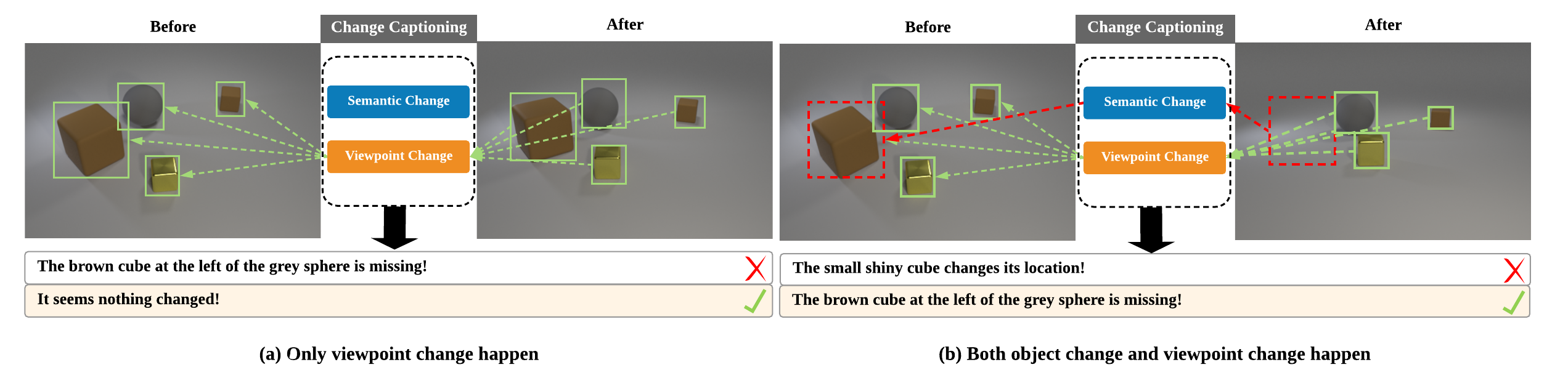}
    \caption{
    Current change captioning methods can be influenced by the viewpoint change between image pairs, which results in generating wrong captions.
    To address this problem, we propose a novel framework that aims at explicitly separating viewpoint changes from semantic changes.
    }
    \label{fig:intro}
\end{figure}

In recent years, a great deal of research has been devoted to \textcolor{\gu}{detecting} changes in various image content~\cite{bruzzone2000automatic,wang2018getnet,alcantarilla2018street,daudt2018fully}, which typically require a model to figure out the differences between the corresponding images. However, most of them focus on the changes in terms of object classes without paying much attention to various attribute changes (e.g., color and size). 
Moreover, \emph{change detection} usually requires pixel-level densely labeled ground-truth data for training the model. Acquiring such labeled data is too time-consuming and labor-intensive, \textcolor{\gu}{some of which, such as} in medical image applications even require human experts for generating annotations.

In addition, current change detection methods usually assume that the differences occur in an ideal situation. However, in practice, changes such as illumination and viewpoint changes occur quite frequently, which \textcolor{\gu}{makes} the change detection task even more challenging.

As humans, we can describe the differences between images accurately, since we can identify the most significant changes in the scene and filter out tiny differences and irrelevant noise. Inspired by the research development from dense object detection to image captioning, which generates natural language descriptions (instead of object labels) to describe the salient information in an image, the \emph{change captioning task} has recently been proposed to describe the salient 
differences between images~\cite{park2019robust,jhamtani2018learning,oluwasanmi2019fully,oluwasanmi2019captionnet}. Arguably, captions describing the changes are more accessible (hence preferred) for users, compared to the map-based labels (e.g., pixel-level binary maps as change labels).

Despite the progress, the existing change captioning methods cannot handle the viewpoint change properly~\cite{park2019robust}. As shown in Fig.~\ref{fig:intro}, the viewpoint change in the images can overwhelm the actual object change leading to incorrect captions. Handling viewpoint changes is more challenging as it requires the model to be agnostic of the changes in viewpoints from different angles, while being sensitive to other salient changes.

Therefore, in this paper, following the prevailing architecture of a visual encoder plus a sentence decoder, we propose a novel viewpoint-agnostic image encoder, called Mirrored Viewpoint-Adapted Matching (\textcolor{\gu}{M-VAM}) encoder, for the change captioning task. Our main idea is to exhaustively measure the feature similarity across different regions in the two images so as to accurately predict the changed and unchanged regions in the feature space. The changed and unchanged regions are formulated as probability maps, which are  used to synthesize the changed and unchanged features. We further propose a \textcolor{\gu}{Reinforcement Attention Fine-tuning (RAF)} process to allow the model to explore other caption choices by perturbing the probability maps.

Overall, the contributions of this work include:
\begin{itemize}
    \item We propose a novel \textcolor{\gu}{M-VAM} encoder that explicitly distinguishes semantic changes from the viewpoint changes by predicting the changed and unchanged regions in the feature space.
    \item We propose a RAF module that helps the model focus on the semantic change regions so as to generate better change captions.
    \item Our model outperforms the state-of-the-art change captioning methods by a large margin in both Spot-the-Diff and CLEVR-Change datasets. Extensive experimental results also show that our method produces more robust prediction results for the cases with viewpoint changes.
\end{itemize}

\section{Related Work}

\subsubsection{Image Captioning.}
Image captioning has been extensively studied in the past few years~\cite{farhadi2010every,vinyals2015show,xu2015show,anderson2018bottom,gu2018unpaired,gu2019unpaired,yang2020deconfounded}. Farhadi~\etal~\cite{farhadi2010every} proposed a template-based alignment method, which matches the visual triplets detected from the images with the language triplets extracted from the sentences. Unlike the template-based solutions, Vinyals~\etal~\cite{vinyals2015show} proposed a Long Short-Term Memory (LSTM) network-based method. It takes the image features extracted from a pre-trained CNN as input and generates the image descriptions word by word. After that, Xu~\etal~\cite{xu2015show} introduced an attention mechanism, which predicts a spatial mask over encoded features at each time step to select the relevant visual information automatically. Anderson~\etal~\cite{anderson2018bottom} attempted to mimic the human's visual system with a top-down attention structure. \textcolor{\gu}{Yang~\etal~\cite{Yang_2019_CVPR} introduced a method to match the scene graphs extracted from the captions and images using Graph Convolutional Networks}.

{Another theme of improvement is to use reinforcement learning (RL) to address the training and testing mismatch (aka. exposure bias) problem  \cite{ranzato2015sequence,yu2017seqgan}. Rennie~\etal~\cite{rennie2017self} proposed the Self-Critical Sequence Training (SCST) method, which optimizes the captioning model with sentence-level rewards. Gu~\etal~\cite{gu2018stack} further improved the SCST-based method and proposed a stack captioning model to generate the image description from coarse to fine.
Chen~\etal~\cite{chen2019improving}, introduced Recurrent Neural Network based discriminator to evaluate the naturality of the generated sentences.}

\subsubsection{Captioning for Change Detection.}
{Change detection tasks aim to identify the differences between a pair of images taken at different time steps to recognize the information behind them. It has a wide range of applications, such as scene understanding, biomedical, and aerial image analysis. Bruzzone~\etal~\cite{bruzzone2000automatic} proposed an unsupervised method to overcome the huge cost of human labeling. Wang~\etal~\cite{wang2018getnet} used CNN to predict the changing regions with their proposed mixed-affinity matrix. Recently, Alcantarilla~\etal~\cite{alcantarilla2018street} proposed a deconvolutional network combined with a multi-sensor fusion SLAM (Simultaneous Localization and Mapping) system to achieve street-view change detection in the real world.}

{Instead of detecting the difference between images, some studies have explored to describe the difference between images with natural language. Jhamtani~\etal~\cite{jhamtani2018learning} first proposed the change captioning task and published a dataset extracted
from the VIRAT dataset~\cite{oh2011large}. They also proposed a language decoding model with an attention mechanism to match the hidden state and features of annotated difference clusters. Oluwasanmi~\etal~\cite{oluwasanmi2019fully} introduced a fully convolutional siamese network and an attention-based language decoder with a modified update function for the memory cell in LSTM. To address the viewpoint change problem in the change captioning task, Park~\etal~\cite{park2019robust} built a synthetic dataset with a viewpoint change between each image pair based on the CLEVR engine. They proposed a Dual Dynamic Attention Model (DUDA) to attend to the images (``before'' and ``after'') for a more accurate change object attention.}

In contrast, in our work, we propose a M-VAM encoder to overcome the \textcolor{\gu}{confusion} raised by the viewpoint changes. Compared with \textcolor{\gu}{Difference Description with Latent Alignment (DDLA)} proposed by Jhamtani~\etal~\cite{jhamtani2018learning}, our method does not need ground-truth difference cluster masks to restrict the attention weights. Instead, our method can self-predict the difference (changed) region maps by feature-level matching without any explicit supervision. Both DUDA~\cite{park2019robust} and \textcolor{\gu}{Fully Convolutional CaptionNet (FCC)}~\cite{oluwasanmi2019fully} apply subtraction between image features at the same location in the given image pair, which has problems with  large viewpoint changes and thus cannot reliably produce correct change captions. Different from them, our model overcomes the  viewpoint change problem by matching objects in the given image pair without the restriction of being at the same feature location.
\section{Method}
\begin{figure}[ht]
    \centering
    \includegraphics[width=0.9\linewidth]{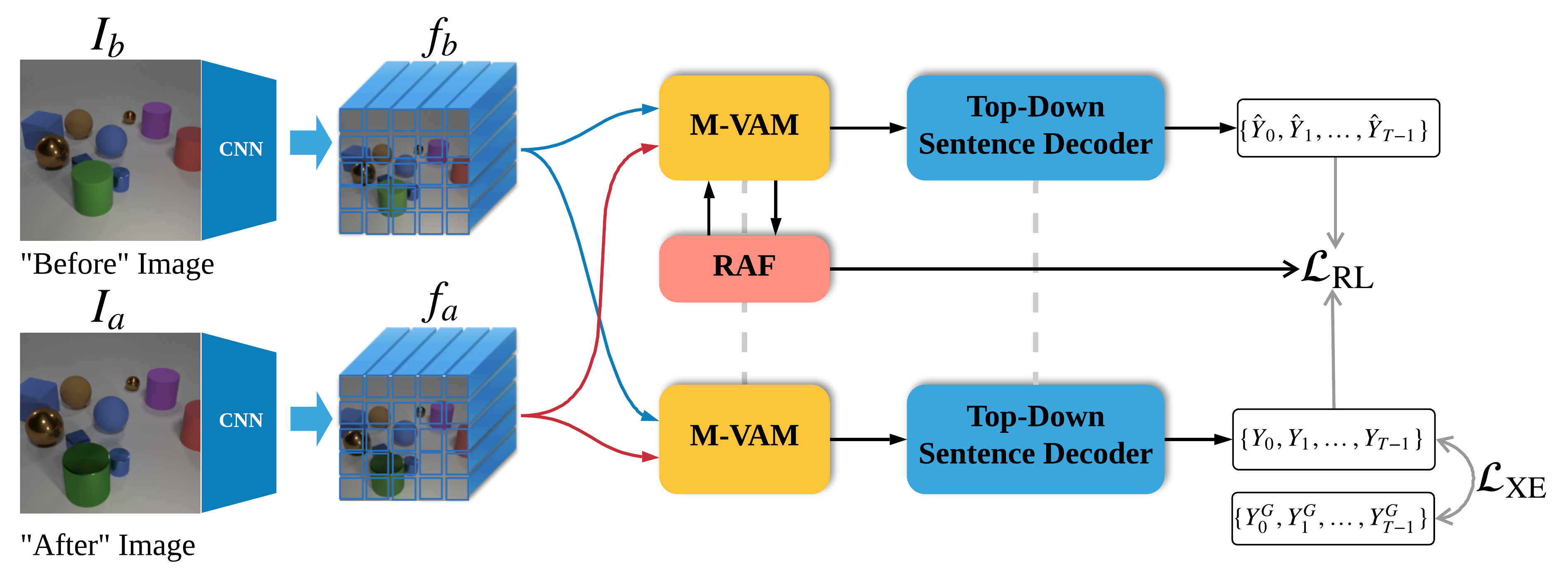}
    \caption{Overview of the proposed change captioning framework. Our model mainly consists of one \textcolor{\gu}{M-VAM} encoder and one top-down sentence decoder. \textcolor{\gu}{We use dashed lines to denote that the parameters of two models are shared with each other}. Note that with a pre-trained CNN, the model first encodes the \textit{before} and \textit{after} images ($I_b$ and $I_a$) into features ($f_b$ and $f_a$) before feeding into M-VAM. The entire model is trained with a cross-entropy loss ($\mathcal{L}_{\text{XE}}$) first, followed by a reinforcement learning loss ($\mathcal{L}_{\text{RL}}$) based on our \textcolor{\gu}{RAF} process. 
    }
    \label{fig:entire}
\end{figure}

Figure~\ref{fig:entire} gives an overview of our proposed change captioning framework. The entire framework basically follows the prevailing encoder-decoder architecture and consists of two main parts: a Mirrored Viewpoint-Adapted Matching encoder to distinguishing the semantic change from the viewpoint-change surrounding, and a top-down sentence decoder to generate the captions.
During training, we use two types of supervisions: we first optimize our model with traditional cross-entropy (XE) loss with the model at the bottom branch. And then we finetune our model with reinforcement learning using both top and bottom branches, which share the parameters, to further explore the information of objects matched in the image pairs.

\subsection{Mirrored Viewpoint-\textcolor{\gu}{Adapted} Matching Encoder}\label{sec:MVAM}
{We first encode the given {``before''} image $I_b$ and the {``after''} image $I_a$ into the spatial image feature maps $f_b\in \mathbb{R}^{W \times H \times C}$ and $f_a\in \mathbb{R}^{W \times H \times C}$, respectively, with a pretrainned CNN, where $W$,  $H$, and $C$ are the width, the height, and the channel size of the feature maps, respectively.}

Fig.~\ref{fig:VAM} shows our proposed M-VAM encoder. It has of two VAM cells with shared parameters. Each VAM cell takes $f_b$ and $f_a$ as input and outputs the changed feature and the unchanged feature. Here, different orders of the inputs lead to different outputs. For example, when a VAM takes the input  
$(f_b, f_a)$, it treats $f_b$ as the reference and computes the changed (or difference) feature $f_{d_{b \rightarrow a}}$ and the unchanged (or background) feature  $f_{bb}$. Similarly, for input order $(f_a, f_b)$, the VAM computes the difference feature $f_{d_{a \rightarrow b}}$ and the background feature $f_{ba}$.

\subsubsection{Viewpoint-\textcolor{\gu}{Adapted} Matching \textcolor{\gu}{(VAM)} Cell}

Given the extracted image features $f_a$ and $f_b$, we aim to detect where the changes are in the feature space. The traditional way is to directly compute the difference of the two feature maps~\cite{park2019robust}, \textcolor{\gu}{which can hardly represent the semantic changes of the two images in the existence of  a viewpoint change}.  
To solve this problem, we propose to identify the changed and the unchanged regions by comparing the similarity between different patches in the two feature maps.

In particular, for any cell location $(x,y)$ in $f_b$ and any cell location $(i,j)$ in $f_a$, we compute their feature-level similarity as follows
 \begin{align}
    \mathrm{E}_{x,y}(i,j) &= f_{b(x,y)}^T f_{a(i,j)}     \label{equ:similar}
 \end{align}

With the obtained $\mathrm{E} \in \mathbb{R}^{WH\times WH}$, we then  generate a synthesized feature map $f_{sb}$ for image $I_b$ from the other image feature $f_a$ as

\begin{align}
    f_{sb(x,y)} &= \sum \limits_{i,j} f_{a(i,j)} \cdot \text{softmax}(\mathrm{E}_{x,y}(i,j))
    \label{recon}
\end{align}
{The softmax in Eq. \eqref{recon} serves as a soft feature selection operation for $f_{a(i,j)}$ according to the similarity matrix $\mathrm{E}$. When an object in $f_{b(x,y)}$ can be found at $f_{a(i,j)}$, the score of $\mathrm{E}_{x,y}(i,j)$ should be higher than the score of other features. In this way, the most corresponding feature can be selected out to synthesize the feature of $f_{sb(x,y)}$, i.e., the objects in the \textit{after} image can ``move back" to the original position in the \textit{before} image at the feature level.}

\begin{figure}[th]
    \centering
    \includegraphics[width=0.9\linewidth]{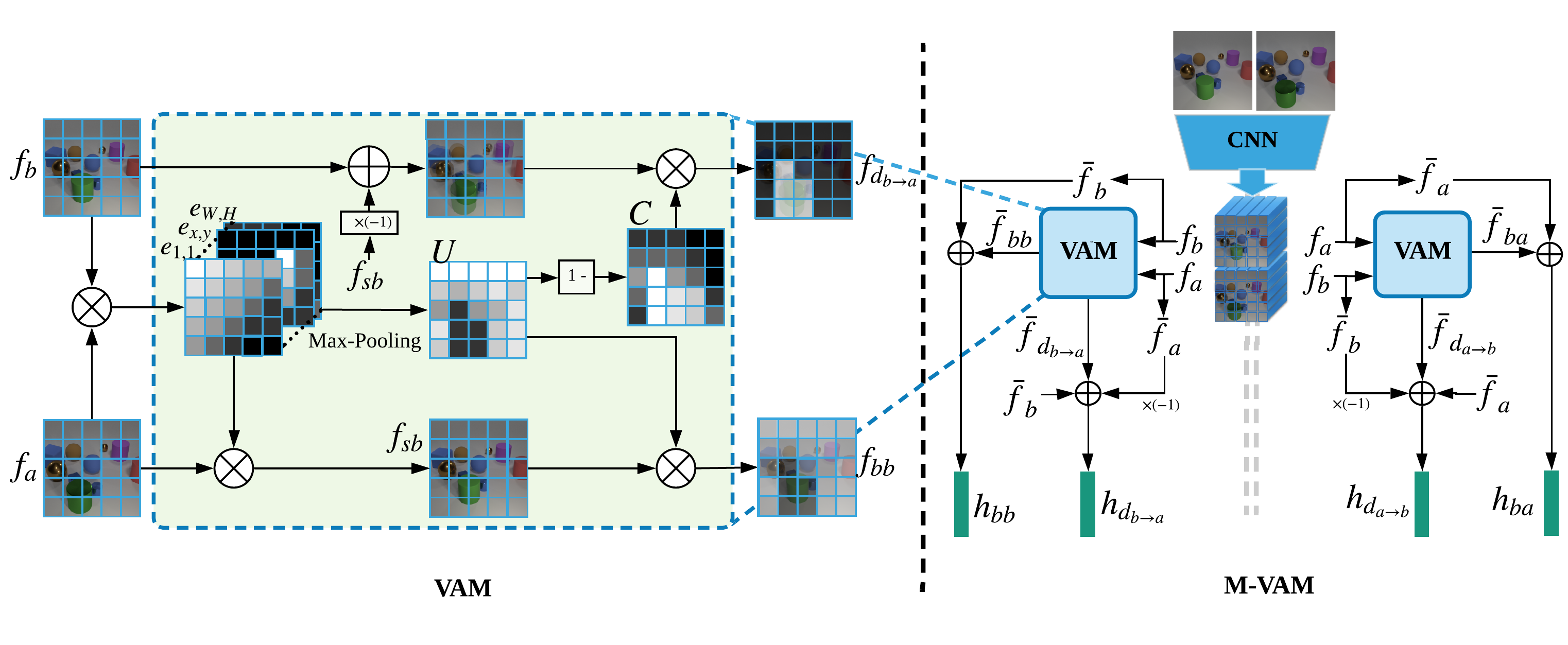}
    \caption{Left: structure of a VAM Cell. Right: proposed M-VAM encoder, which consists of two VAM cells with shared parameters but with the inputs of $f_a$ and $f_b$ in different orders.}

    \label{fig:VAM}
\end{figure}

{So far, our method just blindly searches for the most similar patch in $f_a$ to reconstruct $f_b$ without taking into account the fact that the most similar one in $f_a$ might not be a correct match. Thus, we further compute an unchanged probability map ${U}_{x,y} \in [0,1]^{WH}$ and a changed probability map ${C}_{x,y} \in [0,1]^{WH}$ as follows.
 \begin{align}
    {U}_{x,y} &= \max \limits_{i,j}\left(\sigma (a_u \mathrm{E}_{x,y}(i,j) + b_u)\right)\label{unchange}\\
    {C}_{x,y} &= 1 - {U}_{x,y} \label{equ:C}
\end{align}
where $\sigma$ is the sigmoid function, $a_u$ and $b_u$ are the learned (scalar) parameters.} Finally, as shown in Fig.~\ref{fig:VAM} (left), with the obtained probability maps, the original and the synthesized feature maps $f_b$ and $f_{sb}$, we compute the outputs of the VAM, i.e., the unchanged feature $f_{bb}$ and the changed feature $f_{d_{b \rightarrow a}}$ as: 
\begin{align}
    f_{bb} &= f_{sb} \odot {U} \\
    f_{d_{b \rightarrow a}} &= (f_{b} - f_{sb}) \odot {C} \label{changefeature}
\end{align}
where $\odot$ is the element-wise product.

\subsubsection{\textcolor{\gu}{M-VAM}}

Using one VAM cell with the input order of $(f_b,f_a)$, we can now predict the unchanged and changed features ($f_{bb}$ and $f_{d_{b \rightarrow a}}$), but all of them are with respect to the ``before'' image, $I_b$. For an ideal change captioner, it should catch all the changes in both of the images. Therefore, we introduce the mirrored encoder structure, as shown in Fig.~\ref{fig:VAM} (right). Specifically, we first use the VAM cell to predict the changed and unchanged features w.r.t. $I_b$, and then the same VAM cell is used again to predict the output features w.r.t. $I_a$. Formally, the encoding process can be written as
\begin{align}
    f_{d_{b \rightarrow a}}, f_{bb} &= \text{VAM}(f_b|f_a) \label{equ:mirror_1}\\
    f_{d_{a \rightarrow b}}, f_{ba} &= \text{VAM}(f_a|f_b)\label{equ:mirror_2}
\end{align}

Note that such a mirror structure also implicitly encourages the consistency of the matched patch pairs from the two opposite directions. 

Finally, we treat the features obtained in Eq.~\eqref{equ:mirror_1} and Eq.~\eqref{equ:mirror_2} as the residual information and add them to the original image feature maps:
\begin{align}
    h_{bi} &= \bar{{f}}_{bi} + \bar{{f}}_i, \\
    h_{d_{i \rightarrow j}} &= \bar{{f}}_{d_{i \rightarrow j}} + \bar{{f}}_i - \bar{{f}}_j,
    \label{equ:residual}
\end{align}
where $i, j\in \{a,b\}$ and $i\neq j$, and $\bar{f_*}$ is the resultant feature vector after applying an average pooling on $f_*$.

\subsection{Sentence Decoder}
\label{sec:LD}
\begin{figure}[t!]
    \centering
    \includegraphics[width=0.9\linewidth]{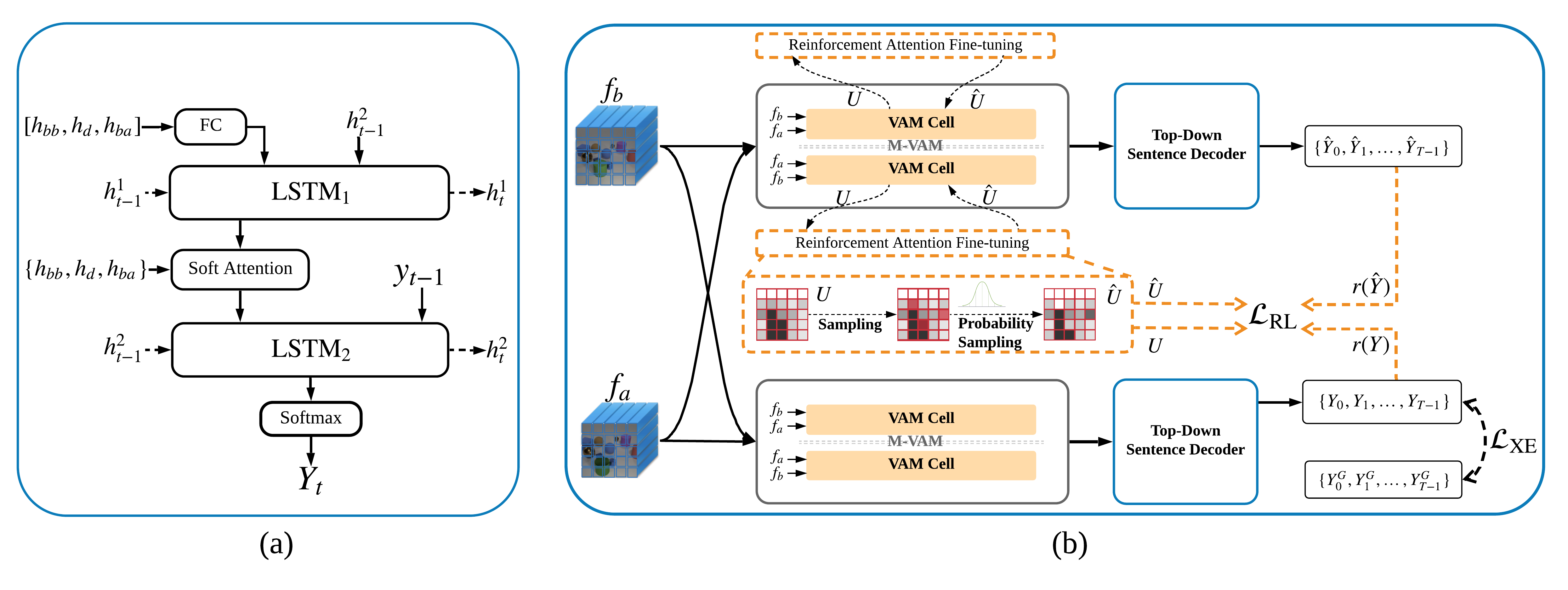}
        \vspace{-0.3in}
\caption{(a): structure of the top-down decoder. (b): proposed RAF process.}
    \label{fig:Dec_RL}
\end{figure}

{As shown in Fig.~\ref{fig:Dec_RL}(a), we adopt a top-down structure~\cite{anderson2018bottom} as our sentence decoder. It  consists of two LSTM networks and a soft attention module. Since we have two changed features generated from the M-VAM encoder, to obtain a consistent changed feature, we first apply a fully connected layer to fuse the two features}
\begin{align}
    h_{d} = \mathbf{W_1}[h_{d_{a \rightarrow b}}, h_{d_{b \rightarrow a}}] + \mathbf{b_1}
\end{align}
{where $\mathbf{W_1}$ and $\mathbf{b_1}$ are learned parameters.}

With the set of the encoded changed and unchanged features $[h_{d},h_{bb},h_{ba}]$, the sentence decoder operates as follows:
\begin{align}
    h^1_t &= \text{LSTM}_1([\mathbf{W_2}[h_{d},h_{bb},h_{ba}] + \mathbf{b_2}, h^2_{t-1}]) \\
    a_{i,t} &= \text{softmax}(\mathbf{W_3} \tanh{(h_i + h^1_t)} + \mathbf{b_3}), \\
    h^2_t &= \text{LSTM}_2([ \sum a_{i,t} \times h_i, y_{t-1}]) \\
    Y_t & \sim \text{softmax}(\mathbf{W_4} h^2_t  + \mathbf{b_4})
    \label{equ:logit}
\end{align}
{where $h^1_t$ and $h^2_t$ are the hidden states of the first LSTM and the second LSTM, respectively, $a_{i,t}$ is the soft attention weight for the visual feature $h_i$ at time step $t$, $i \in \{d, bb, ba\}$ refers to different input features, $y_{t-1}$ is the previous word embedding, and $Y_t$ is the output word drawn from the dictionary at time step $t$ according to the softmax probability.}

\subsection{Learning Process}
\subsubsection{Cross-entropy Loss}
{We first train our model by minimizing the traditional cross-entropy loss. It aims to  maximize the probability of the ground truth caption words given the two images:}
\begin{align}
    \mathcal{L}_{\text{XE}} = - \sum_{t=0}^{T-1} \log(p_{\theta}(Y^G_t|Y^G_{0:t-1}, I_a, I_b))\label{eq:xe_loss}
\end{align}
where $\theta$ is the parameter of our model, and $Y^G_t$ is the $t$-th ground-truth word.

\subsubsection{Reinforcement Attention Fine-tuning (RAF)} Training with the XE loss alone is usually insufficient, and the model suffers from the exposure bias problem \cite{bengio2015scheduled}. The ground-truth captions of an image contain not only language style information but also information about preferential choices. When it comes to describing the change in a pair of images, a good captioning model needs to consider the relations between the objects and how the objects change. In other words, the model needs to learn what and when to describe. Besides, as \textcolor{\gu}{humans,} when we describe the changed objects, we often include some surrounding information. All these different expressions are related to different visual information in the images in a complicated way. To model such complicated relations, we propose to predict attention regions (i.e., focus areas to describe the changes) in a more human-like way that can help the model to generate more informative captions. 
Different from image captioning tasks, the images here are encoded only once to decode the sentence 
(i.e., before staring the decoding, our M-VAM encoder encodes an image pair into features, $h_d, h_{bb}, h_{ba}$, and keeps them unchanged during the entire decoding process as opposed to computing a different attention or context vector at each decoding step), for which the standard RL would have much larger influence on the decoder rather than on the encoder. Therefore, we propose a RAF process based on the attention perturbation to further optimize our encoder-decoder model.
 
As shown in Fig.~\ref{fig:Dec_RL}(b), our RAF process generates two captions for the same image pair. The first caption $Y = \{Y_0, Y_1, \cdots, Y_{T-1} \}$ is generated in the usual way using our model as shown in the bottom branch of Fig.~\ref{fig:Dec_RL}(b), while the second caption $\hat{Y} = \{\hat{Y}_0, \hat{Y}_1, \cdots, \hat{Y}_{T-1} \}$ is generated in a slightly different way according to a sampled probability map $\hat{{U}}$ (and $\hat{{C}}$), as shown in the top branch. Here, $\hat{{U}}$ is a slightly modified map of $U$ by introducing some small noise, which will lead to the generation of a different caption. By comparing the two captions ${Y}$ and $\hat{Y}$, and rewarding the better one with the reinforcement learning, our model can  essentially  \emph{explore} for a better solution.

Specially, to generate $\hat{{U}}$, we first randomly select some elements to perturb from the unchanged probability map ${U}$ with a sampling rate of $\gamma$. For the selected element ${{U}}_{i,j}$, we get its perturbed version by sampling from the following Gaussian distribution:
\begin{align}
    \hat{U}_{i,j} \sim \mathcal{N}(U_{i,j}, c)
    \label{equ:sample}
\end{align}
where the standard deviation $c$ is a small number (e.g., $0.1$ in our experiments) that is considered as a  hyperparameter. For the elements that are not selected for perturbation, $\hat{U}_{i,j}=U_{i,j}$. With $\hat{{U}}$, we can then calculate the corresponding changed map $\hat{{C}}$ using Eq. \eqref{equ:C} and generate the caption ${\hat{Y}}$ accordingly.

After that, we calculate the CIDEr scores for the two captions and treat them as rewards, i.e.,  $r({Y}) = \text{CIDEr}(Y)$ and $r(\hat{{Y}}) = \text{CIDEr}(\hat{Y})$. If $r(\hat{{Y}})$ is higher than  $r({Y})$, it means that $\hat{Y}$ describes the change of the objects better than ${Y}$ (according to CIDEr). Consequently, the predicted probability map ${U}$ should be close to the sampled probability map $\hat{{U}}$. On other other hand, if $r(\hat{{Y}})$ is smaller than  $r({Y})$, it suggests that the sampled probability map is worse than the original one, thus should not be used to guide the learning.  Based on this intuition, we define our reinforcement loss as follows.
\begin{align}
    \mathcal{L}_{\text{RL}} = \mathbbm{1}^+ (r(\hat{{Y}}) - r({Y})) ||\hat{{U}} - {U}||^2_2,
    \label{equ:rlloss}
\end{align}
{where $\mathbbm{1}^+(x)$ represents $\max (x, 0)$.}
{The gradients from $\mathcal{L}_{\text{RL}}$ will backward to the M-VAM module. The overall training loss becomes}:
\begin{align}
    \mathcal{L}_{\text{XE+RL}} = \lambda \mathcal{L}_{\text{RL}} + \mathcal{L}_{\text{XE}}
    \label{equ:mixloss}
\end{align}
where $\lambda$ is a hyper-parameter to control the relative weights of the two losses.

\section{Experiment}
\subsection{Datasets and Metrics}

{We evaluate our method on two published datasets: Spot-the-Diff~\cite{jhamtani2018learning} and CLEVR-Change~\cite{park2019robust}. {Spot-the-Diff} contains  13,192 image pairs extracted from the VIRAT dataset~\cite{oh2011large}, which were captured by stationary ground cameras. 
Each image pair can differ in multiple ways; on average, each image pair contains 1.86 reported differences. 
{CLEVR-Change} is a synthetic dataset with a set of basic geometry objects, which is generated based on the CLEVR engine. Different from Spot-the-Diff dataset, to address the viewpoint change problem, CLEVR-Change involves a viewpoint change between each image pair. The dataset contains 79,606 image pairs and 493,735 captions. 
The changes in the dataset can be categorized into six cases, which are Color, Texture, Movement, Add, Drop, and Distractor.}

{We evaluate our generated captions with five evaluation metrics: BLEU~\cite{papineni2002bleu}, METEOR~\cite{denkowski2014meteor}, CIDEr~\cite{Vedantam2015CIDEr}, ROUGE$\_$L~\cite{lin2004rouge}, and SPICE~\cite{anderson2016spice}. For comparison in terms of BLEU scores, we mainly show the results for BLEU4 which reflects the matching ratio of 4-word sub-sequences between the generated caption and the ground truth caption.}

\subsection{Training Details}

We use ResNet101~\cite{he2016deep} trained on the ImageNet dataset~\cite{krizhevsky2012imagenet} to extract the image features.
The images are resized to $224 \times 224$ before extracting the features. The feature map size is $\mathbb{R}^{14 \times 14 \times 1024}$.
During the training, we set the maximum length of captions as 20 words. 
The dimension of hidden states in the Top-Down decoder is set to 512. We apply a drop-out rate of 0.5 before the word probability prediction (Eq.\eqref{equ:logit}). We apply ADAM~\cite{kingma2014adam} optimizer for training.

When training with the cross-entropy loss, we train the model for 40 epochs with a learning rate of 0.0002. During the reinforcement attention fine-tuning (RAF) process, we set the learning rate to 0.00002. Due to the low loss value from Eq.~\eqref{equ:rlloss}, we set the hyperparameter $\lambda$ in Eq.~\eqref{equ:mixloss} as 1000.0 to put more weight to the reinforcement learning loss. In the RAF process, we set the random sampling rate $\gamma$ as 0.025. 
As for the Gaussian distribution (Eq.~\eqref{equ:sample}), we set $c$ as 0.1.

\subsection{Model Variations}
To analyze the contribution of each component in our model, we consider the following two variants of our model.

    \begin{itemize} 
    
    \item[1)] \textbf{M-VAM}: This variant   models the encoder-decoder captioning framework with the proposed M-VAM encoder introduced in \textcolor{\gu}{Sec.~\ref{sec:MVAM}} and the top-down decoder introduced in \textcolor{\gu}{Sec.~\ref{sec:LD}.}
    \item[2)] \textbf{M-VAM + RAF}: This variant fine-tunes M-VAM with the proposed RAF process. 
\end{itemize}

\subsection{Results}

\subsubsection{Quantitative Results on {Spot-the-Diff}.}  Quantitative results of different methods on {Spot-the-Diff} dataset are shown at Table~\ref{tab:StD}. We compare with the four existing methods -- FCC~\cite{oluwasanmi2019fully},  SDCM~\cite{oluwasanmi2019captionnet}, DDLA~\cite{jhamtani2018learning} and DUDA~\cite{park2019robust}. It can be seen that both of our models outperform existing methods in all evaluation measures. Compared with FCC, \textcolor{\gu}{M-VAM} provides improvements of 0.2, 1.4 and 1.3 in BLEU4, ROUGE\_L and CIDEr, respectively. After using the RAF process, the scores are further improved from 10.1 to 11.1 in BLEU4, 31.3 to 33.2 in ROUGE\_L,  38.1 to 42.5 in CIDEr, and 14.0 to 17.1 in SPICE. The big gains in CIDEr (+4.4) and SPICE (+3.1) for the \textcolor{\gu}{RAF} fine-tuning demonstrate its effectiveness in change captioning.

\begin{table}[ht]
    \begin{center}
        \small
        \caption{Comparisons of the results of different methods on {Spot-the-Diff} dataset.}
        \label{tab:StD}
        
        \begin{tabular}{l|p{1.8cm}p{1.8cm}p{1.8cm}p{1.8cm}p{1.8cm}p{1.8cm}p{1.8cm}} 
        
            \hline
            Model    & BLEU4 & ROUGE\_L    & METEOR         & CIDEr  &SPICE\\
            \hline
            DUDA~\cite{park2019robust} &8.1 &28.3 &11.5 & 34.0 &-\\ 
            FCC~\cite{oluwasanmi2019fully} &9.9 &29.9 &\textbf{12.9} &36.8 &-\\
            SDCM~\cite{oluwasanmi2019captionnet} &9.8 &29.7 &12.7 &36.3 &-\\
            DDLA~\cite{jhamtani2018learning} &8.5 & 28.6 &12.0 &32.8 &-\\
            \hline
            M-VAM &10.1 & 31.3 &12.4 &38.1 &14.0\\
            M-VAM + RAF &\textbf{11.1} &\textbf{33.2} &\textbf{12.9} &\textbf{42.5} &\textbf{17.1}\\
            \hline
        \end{tabular}
    \end{center}
\end{table}

\begin{table}[ht]
    \begin{center}
        \small
        \caption{Comparisons of the results of different methods on {CLEVR-Change} dataset.}
        \label{tab:CLEVR}
        
        \begin{tabular}{l|p{1.8cm}p{1.8cm}p{1.8cm}p{1.8cm}p{1.8cm}p{1.8cm}p{1.8cm}} 
        
            \hline
            Model    & BLEU4 & ROUGE\_L    & METEOR         & CIDEr  &SPICE\\
            \hline
            DUDA~\cite{park2019robust} &47.3 &- &33.9 & 112.3 & 24.5\\ 
            \hline
            \multicolumn{6}{c}{\textit{{With Distractor}}}\\
            \hline
            M-VAM &50.3 & 69.7 &37.0 &114.9 &30.5\\
            M-VAM + RAF &\textbf{51.3} &70.4 &\textbf{37.8} &115.8 & 30.7\\
            \hline
            \multicolumn{6}{c}{\textit{{Without Distractor}}}\\
            \hline
            M-VAM &45.5 & 69.5 &35.5 &117.4 &29.4\\
            M-VAM + RAF &50.1 &\textbf{71.0} &36.9 &\textbf{119.1} &\textbf{31.2}\\
            \hline
        \end{tabular}
    \end{center}
\end{table}

\subsubsection{Quantitative Results on {CLEVR-Change}.} In Table~\ref{tab:CLEVR}, we compare our results with DUDA~\cite{park2019robust} on the CLEVR-Change dataset. {CLEVR-Change} contains additional distractor image pairs, which may not be available in other scenarios. Therefore, we performed separate experiments with and without the distractors to show the efficacy of our method.

We can see that our model tested on the entire dataset (with distractor) achieves state-of-the-art performance and outperforms DUDA by a wide margin. Our M-VAM pushes the BLEU4, METEOR, SPICE and CIDEr scores from 47.3, 33.9, 24.4 and 112.3 to  50.3, 69.7, 37.0, 30.7 and 114.9, respectively. The RAF process further improves the scores by about 1 point in all measures.

Our model also achieves impressive performance on the dataset without the distractors. The M-VAM achieves a CIDEr score of 117.4. RAF further improves the scores -- from 45.5 to 50.1 in BLEU4 and from 117.4 to 119.1 in CIDEr. The gains are higher in this setting than the gains with the entire dataset. There is also a visible increase in SPICE compared with the models trained with the distractors (from 30.7 to 31.2). This is because, without the influence of the distractors, the model can distinguish the semantic change more accurately from the viewpoint change.

\begin{table}[t!]
    \begin{center}
        \small
        \caption{The Detailed breakdown of evaluation on the {CLEVR-Change} dataset.}
        \label{tab:detailed}
        
        \begin{tabular}{l|l|p{1cm}p{1cm}p{1cm}p{1cm}p{1cm}p{1cm}p{1cm}|}
        
            \hline
            Model & Metrics   & C & T & A & D & M & DI  \\
            \hline
            DUDA~\cite{park2019robust}          &CIDEr &120.4 &86.7 &108.2 &103.0 &56.4 &110.8\\ 
            M-VAM + RAF (w DI)   &CIDEr &122.1 &98.7 &126.3 &115.8 &82.0 &\textbf{122.6}\\
            M-VAM + RAF (w/o DI) & CIDEr&\textbf{135.4} &\textbf{108.3} &\textbf{130.5} &\textbf{113.7} &\textbf{107.8} &-\\
            \hline
            DUDA~\cite{park2019robust}          &METEOR &32.8 &27.3 &33.4 &31.4 &23.5 &45.2\\ 
            M-VAM + RAF(w DI)   &METEOR &35.8 &32.3 &37.8 &36.2 &27.9 &\textbf{66.4}\\
            M-VAM + RAF (w/o DI) &METEOR &\textbf{38.3} &\textbf{36.0} &\textbf{38.1} &\textbf{37.1} &\textbf{36.2} &-\\
            \hline
            DUDA~\cite{park2019robust}          &SPICE &21.2 &18.3 &22.4 &22.2 &15.4 &28.7\\ 
            M-VAM + RAF (w DI)   &SPICE &28.0 &26.7 &30.8 &32.3 &22.5 &\textbf{33.4}\\
            M-VAM + RAF (w/o DI) &SPICE &\textbf{29.7} &\textbf{29.9} &\textbf{32.2} &\textbf{33.3} &\textbf{30.7} &-\\
            \hline
        \end{tabular}
    \end{center}
\end{table}

Table \ref{tab:detailed} shows the detailed breakdown of the evaluation according to the categories defined in~\cite{park2019robust}. In the table, ``Metrics" represent the metric used to evaluate the results for a certain model. ``C" represents the changes related to ``Color" attributes. Similarly, ``T", ``A", ``D", ``M" and ``DI" represent ``Texture", ``Add", ``Drop", ``Move" and ``Distractor", respectively.

Comparing our results with DUDA, we see that our method can generate more accurate captions for all kinds of changes. In particular, it obtains large improvements in CIDEr for ``Move"  and ``Add" cases (from 108.2 to 126.3 and from 56.4 to 82.0, respectively). For SPICE, the ``Drop" cases have the largest improvement (from 22.2 to 32.3) among the six cases. This indicates that our method can better distinguish the object movement from a viewpoint change scene, which is the most challenging case as reported in \cite{park2019robust}. The model can also match the objects well when objects are missing from the `before' image or added to the `after' image. Referring to the METEOR score, we get the highest improvement for the  ``Distractor" cases. This means that our method can better discriminate the scene only with the viewpoint change from the other cases.

Comparing the model trained with and without the distractors, we can see that the performance is further boosted in ``Color", ``Texture" and ``Move" cases. The SPICE scores also show an increase in the ``Move" case (from 22.5 to 30.7). It means that such cases are easier to be influenced by the distractor data. This is because the movement of objects can be confused with the viewpoint movement. As for ``Color" and ``Texture", it may be caused by the surrounding description mismatching according to the observation. When the model is trained without the distractors, our RAF process, which is proposed to capture a better semantic surrounding information, is more efficient and thus improves the performance in ``Color" and ``Texture".

\begin{figure}[t!]
    \centering
    \includegraphics[width=0.95\linewidth]{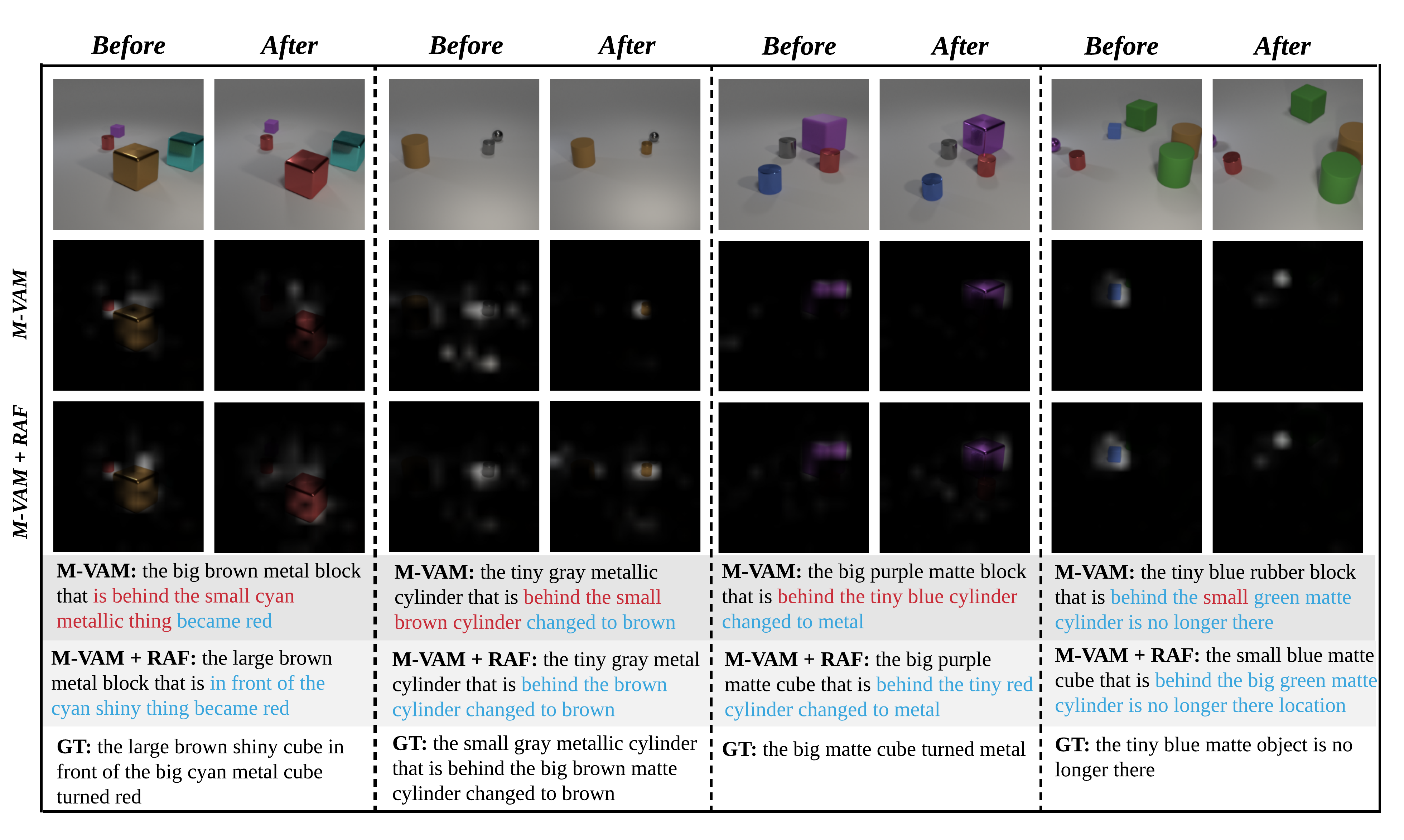}
    \caption{Visualization of examples from the CLEVR-Change dataset. Red and blue color respectively denote the inconsistent and consistent words. The middle two rows show the changed probability maps of the two models.}
    \label{fig:visual}
\end{figure}

\begin{figure}[t!]
    \centering
    \includegraphics[width=0.95\linewidth]{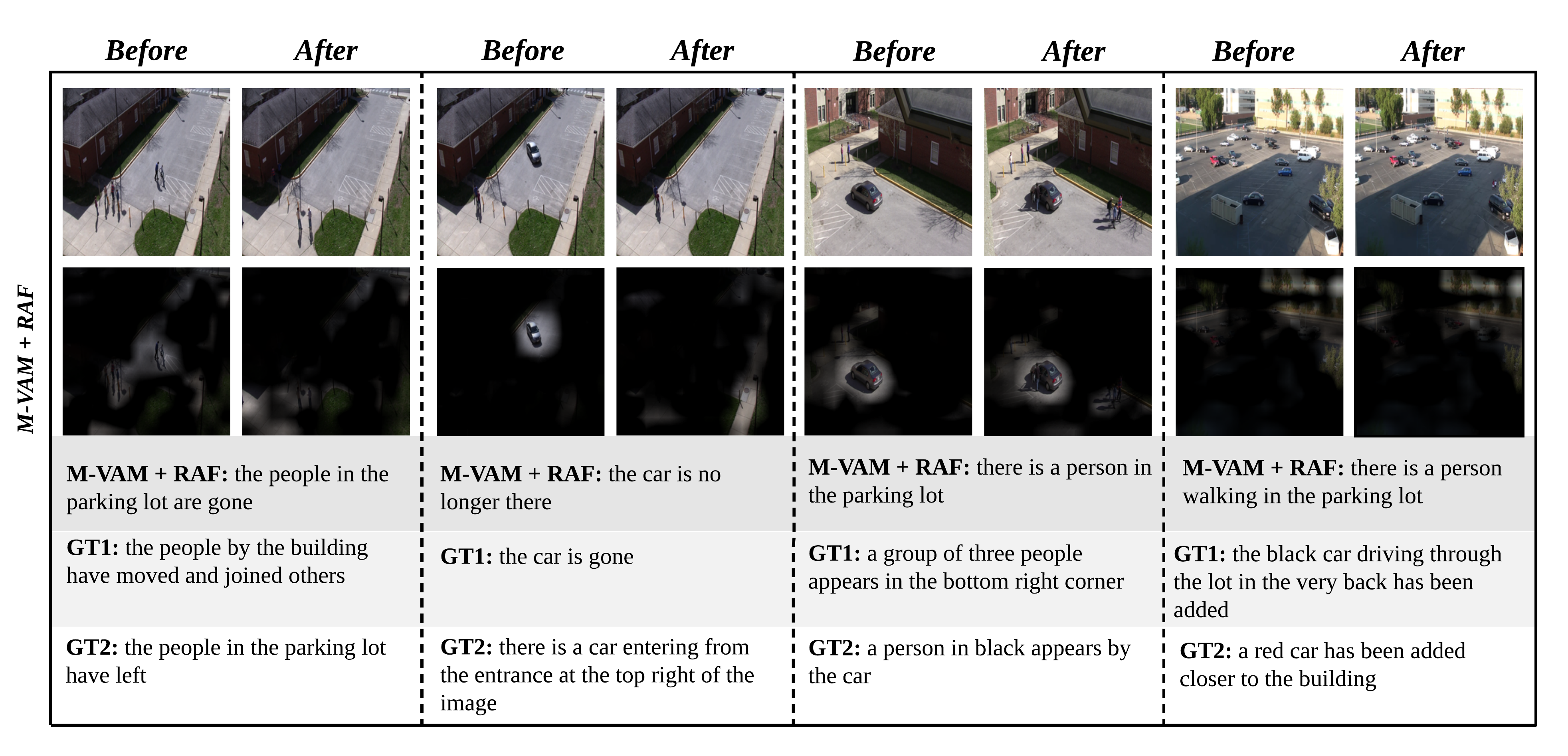}
    \caption{\textcolor{\gu}{Visualizations of the generated captions on Spot-the-Diff.}}
    \label{fig:visual_StD}
\end{figure}

\subsubsection{Qualitative Results} We show the visualization of some examples from the {CLEVR-Change} dataset  in Fig.~\ref{fig:visual}. The images at the top of the figure are the given image pairs. The images in the middle two rows are the changed probability maps of  \textcolor{\gu}{M-VAM} and \textcolor{\gu}{M-VAM+RAF}, respectively. Comparing the probability maps, we can see that the maps after the \textcolor{\gu}{RAF} process are brighter and more concentrated at the changed regions than the ones from \textcolor{\gu}{M-VAM}, which indicates \textcolor{\gu}{RAF} can force the model to focus on the actual changed regions. 
Note that a rigorous overlap comparison is not meaningful here since the model does not perform an explicit change detection to find the objects in the pixel level (the size of the attention map is only $14 \times 14$).

In the first example, we can see that the attention map from \textcolor{\gu}{M-VAM} covers the changed object in the image only partially, while also attending to some irrelevant portions above the changed object. Thus, with the shifted attention region in the ``after" image, the model  mistakenly infers that the cube is behind (as opposed to in the front) the cyan cube. {The \textcolor{\gu}{RAF} fine-tuning helps the model focus more appropriately and model the change correctly.} In the second column, we observe that \textcolor{\gu}{RAF} can  disregard small irrelevant attentions and keep concentration on the semantic change that matters (e.g., the brown cylinder in the ``after" image in the second column). In the third column, the example shows that our model can detect changes in \emph{Texture} successfully, which has a low reported score in Park \etal~\cite{park2019robust}. In the last column, we can see that our model shows robustness to large viewpoint change; it can predict a clear and accurate changed probability map, filtering the distraction caused by the viewpoint shift. The small bright region in the ``after" image represents a location where the blue cube is missing.

When summarizing an image with a short description like caption, humans tend to describe the surroundings and relationships with respect to the central entities. In the third column, the M-VAM picks ``tiny blue cylinder" to describe the change of ``big purple matte cube'', whereas the M-VAM+RAF picks ``tiny red cylinder" which is closer to the central entity.

Fig.~\ref{fig:visual_StD} shows examples from the {Spot-the-Diff} dataset. From the first to the third column, we can see  that our model is also good at finding the difference in real-world scenes. In the first example, our model could successfully detect the people and their non-existence.  In the second example, our model can detect the car and the fact that it is not in its position  anymore. Although there is a mismatch at the changed probability map of the ``after" image, the information from the `before' image is strong enough to filter such noise and generate an accurate caption. Similarly, our model can successfully detect the changes between the images in the third example and focus on the most salient information (i.e., the appearance of a person near the car) to generate the caption somewhat correctly.

A negative sample is shown in the fourth column. We suspect this is because of the resize operation,due to which some objects become too small to be recognized. As a result,  although the matching map is able to find all the changes, the semantic change information is overshadowed by the information in the surrounding, leading to a misunderstanding.

\section{Conclusion}
In this paper, we have introduced the novel Mirrored Viewpoint-Adapted Matching encoder for the change captioning task to distinguish semantic changes from viewpoint changes. Compared with other methods, our M-VAM encoder can accurately filter out the viewpoint influence and figure out the semantic changes from the images. To further utilize the language information to guide the difference judgment, we have further proposed a Reinforcement Attention Fine-tuning process to supervise the matching map with the language evaluation rewards. Our proposed RAF process can 
highlight the semantic difference in the changed map. The test results have shown that our proposed method surpasses all the current methods in both Spot-the-Diff and CLEVR-Change datasets.

\textbf{Acknowledgement} This research is partially supported by the MOE Tier-1 research grants: RG28/18 (S) and the Monash University FIT Start-up Grant.


\end{document}